\documentclass{article}
\usepackage[utf8]{inputenc}
\usepackage{lipsum}
\usepackage{natbib}
\usepackage{caption}
\usepackage{graphicx}
\usepackage{cite}
\usepackage{graphicx,wrapfig,lipsum}
\usepackage{subcaption}
\usepackage{amsmath}
\usepackage{wrapfig}
\usepackage{hyperref}
\usepackage{algorithm}
\usepackage{algpseudocode}

\hypersetup{
    colorlinks=false,
    linkcolor=blue,
    filecolor=magenta,      
    urlcolor=cyan,
    pdfpagemode=FullScreen,
    }
\usepackage[table,svgnames]{xcolor} % provides the \rowcolors command 
\usepackage{caption} % for improved spacing around the caption

\usepackage{array} % enables >{...} in the coumn specifier section, used in table 2 & 3
\usepackage{booktabs} % for improved spacing around horizontal lines, used in example 3, incompatible with vertical lines, be careful if you want to to combine it with color

\usepackage[column=0]{cellspace} % for adding a small amount of space above and below each cell, only used in table 2
\usepackage{geometry}
 \usepackage{listings}
\usepackage{xcolor}

\definecolor{codegreen}{rgb}{0,0.6,0}
\definecolor{codegray}{rgb}{0.5,0.5,0.5}
\definecolor{codepurple}{rgb}{0.58,0,0.82}
\definecolor{backcolour}{rgb}{0.95,0.95,0.92}

\lstdefinestyle{mystyle}{
    backgroundcolor=\color{backcolour},   
    commentstyle=\color{codegreen},
    keywordstyle=\color{magenta},
    numberstyle=\tiny\color{codegray},
    stringstyle=\color{codepurple},
    basicstyle=\ttfamily\footnotesize,
    breakatwhitespace=false,         
    breaklines=true,                 
    captionpos=b,                    
    keepspaces=true,                 
    numbers=left,                    
    numbersep=5pt,                  
    showspaces=false,                
    showstringspaces=false,
    showtabs=false,                  
    tabsize=2
}

\title{\bfseries Rice Paddy Disease Classification Using CNNs}
\author{Charles O'Neill}
\date{July 2022}

\begin{document}

\maketitle

\abstract{Rice is a staple food in the world's diet, and yet huge percentages of crop yields are lost each year to disease. To combat this problem, people have been searching for ways to automate disease diagnosis. Here, we extend on previous modelling work by analysing how disease-classification accuracy is sensitive to both model architecture and common computer vision techniques. In doing so, we maximise accuracy whilst working in the constraints of smaller model sizes, minimum GPUs and shorter training times. Whilst previous state-of-the-art models had 93\% accuracy only predicting 5 diseases, we improve this to 98.7\% using 10 disease classes.}

\section{Introduction}
\begin{figure}
     \centering
     \begin{subfigure}[b]{0.16\textwidth}
         \centering
         \includegraphics[width=\textwidth]{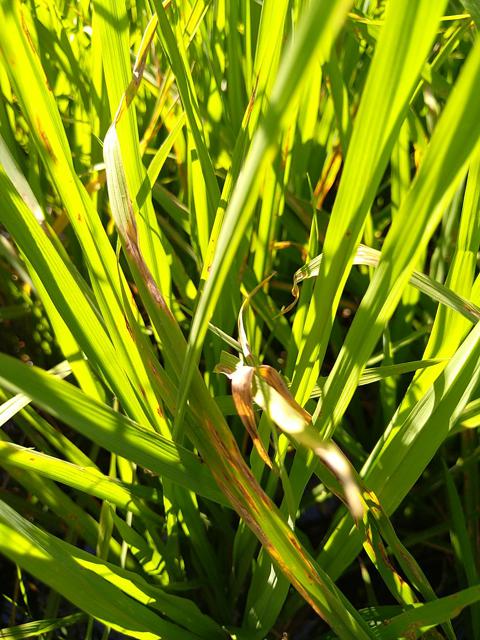}
         \caption{Blast}
         \label{fig:blast}
     \end{subfigure}
     \hspace{6mm}
     \begin{subfigure}[b]{0.16\textwidth}
         \centering
         \includegraphics[width=\textwidth]{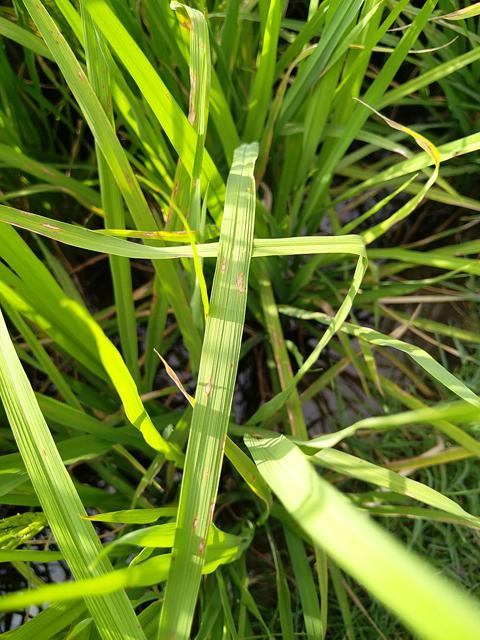}
         \caption{Brown spot}
         \label{fig:brown_spot}
     \end{subfigure}
     \hspace{6mm}
     \begin{subfigure}[b]{0.16\textwidth}
         \centering
         \includegraphics[width=\textwidth]{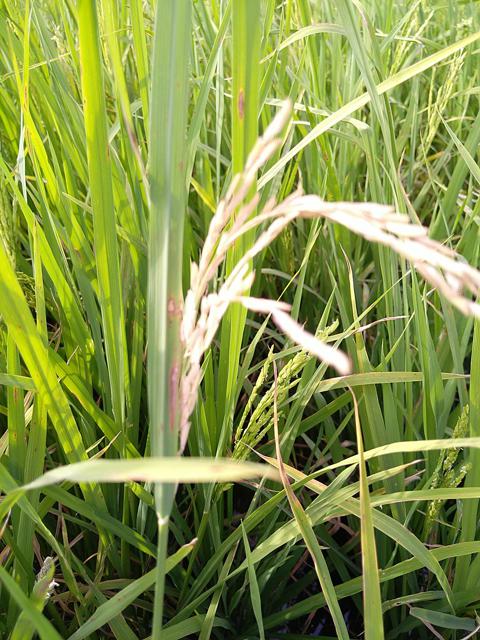}
         \caption{Dead-heart}
         \label{fig:dead_heart}
     \end{subfigure}
        \caption{Three common rice paddy diseases.}
        \label{fig:diseases}
\end{figure}

Rice (\textit{Oryza sativa}) is one of the most important food crops in the world today. It is considered a staple component in the diet of over half the world population \citep{calpe2002rice}. However, 70\% of crop yields are lost each year to common diseases \citep{xu2017uorf}, such as those shown in Figure \ref{fig:diseases}. Because of this, prevention of common rice crop diseases is critical due to food security concerns \citep{fina2013automatic}. Whilst most rice disease diagnosis is done manually by local experts \citep{sethy2020detection}, this is time-consuming, unreliable and often inaccessible. This engenders a research \textbf{question}: how can we use mathematical tools and images to accurately and efficiently diagnose rice paddy disease?

As such, many groups have attempted to use mathematical modelling to perform algorithmic rice paddy disease diagnosis. Support vector machines have previously been used to detect diseased leaves, albeit without classifying them \citep{singh2017detection}. \citet{islam2018faster} developed on this work to take an RGB sample of the affected portion of the leaf, and use this as input to a naïve Bayes as a disease classifier. Machine learning technology has even been layered on top of thermal imaging cameras to identify disease by temperature \citep{zhu2018application}. 

However, these methods were limited, as they relied on manual feature extraction (often requiring an expert) as well as often expensive technology. As a result, several groups have recently attempted to apply deep learning to the problem. For instance, \citet{deng2021automatic} achieved 91\% classification accuracy on dataset highly similar to ours. Using a VGGNet with an Inception Module, \citet{chen2020using} achieved an accuracy of 92\%. \citet{rahman2020identification} improved on this with a standard convolutional neural network (CNN) architecture, reaching 93.3\% accuracy. Others have even designed systems for real-time disease diagnosis using mobile technology \citep{temniranrat2021system}. Admittedly, some of this success may be attributed to the number of classes predicted: for instance, \citet{chen2020using} only attempted to predict five diseases.

Obviously, the central goal of any model is performance: how accurately does the model classify the diseases? This analysis will examine a number of different techniques for maximising accuracy. In addition to this, however, we consider how to optimise this performance under the constraints of limited model size, memory, and GPU speed. This is a key consideration in many aspects of model development, particularly deep learning. For instance, farmers attempting to train their own models on specific crops are unlikely to have access to world-class GPUs and unlimited memory. Thus, the \textbf{main contribution} of this research is how model performance varies within the constraints outlined above. We \textbf{hypothesise} that we can develop significantly more accurate models with less compute power and shorter training times.

\subsection{Data}
The data were collected by the Paddy Doctor project team at the Department of Computer Science and Engineering, Manonmaniam Sundaranar University, Tirunelveli, India. It consists of a base dataset of 10,407 labelled paddy leaf images, with ten classes of disease. We use these images for both training and validation. An additional holdout test set of 3,469 images are reserved for further evaluating the model. The data are available \href{https://www.kaggle.com/competitions/paddy-disease-classification/data}{here}.

\section{Model}
Computer vision has seen big advancements in a relatively short time-period by leveraging the power of deep learning, particularly convolutional neural networks (CNNs). Such models use backpropagation to update kernels which perform convolution operations, maintaining sparsity and preserving the spatial integrity of input images \citep{shorten2019survey}.

\begin{figure}
     \centering
     \begin{subfigure}[b]{0.16\textwidth}
         \centering
         \includegraphics[width=\textwidth]{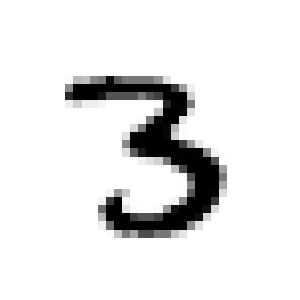}
         \caption{Original image}
         \label{fig:blast}
     \end{subfigure}
     \hspace{6mm}
     \begin{subfigure}[b]{0.16\textwidth}
         \centering
         \includegraphics[width=\textwidth]{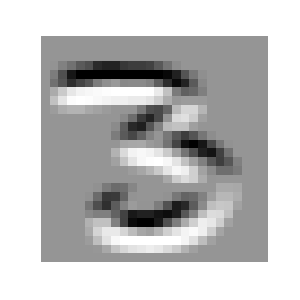}
         \caption{Top-edge kernel}
         \label{fig:top-edge}
     \end{subfigure}
     \hspace{6mm}
     \begin{subfigure}[b]{0.16\textwidth}
         \centering
         \includegraphics[width=\textwidth]{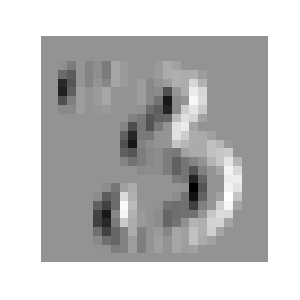}
         \caption{Left-edge kernel}
         \label{fig:left-edge}
     \end{subfigure}
        \caption{Different kernels extract different features through convolution.}
        \label{fig:convolution}
\end{figure}

A CNN can be decomposed into four operations: convolution, non-linearity, pooling, and output. As the name suggests, the key idea of CNNs is the convolution operation. This is motivated by how humans detect images through the visual cortex; we look for sharp edges and gradients to distinguish objects and recognise depth. Since images are essentially 2D-matrices, we can use what amounts to element-wise matrix multiplication (Hadamard product) over this matrix to extract features. Depending on the matrix we use for convolution, we can extract vertical edges, horizontal edges and many other types of visual patterns. Regardless, the aim of the convolution operator (kernel) is to reduce the image size across all three RBG dimensions by sliding the kernel over the image. Figure \ref{fig:convolution} shows the effects of different convolutions, highlighting how different elements in the kernel can detect different features.

The key innovation of CNNs (introduced with Lenet by \citet{bottou1994comparison}) was to not manually choose the kernels used for feature extraction, but rather treat the weights of the kernels as parameters that can be optimised via backpropagation in a neural network. In this way, deep learning finds the best feature detectors that create the richest (and simplest) representation of the image that can be passed to a fully-connected (FC) layer for output prediction. Along with steps such as pooling (which takes neighbourhood-based averages of elements in the image matrix to further reduce dimension), an extremely effective computer vision model can be constructed from scratch in a deep learning framework such as Pytorch with minimal code. Thus, we select CNN as the architecture for our disease-classifying model.

\begin{figure}
    \centering
    \includegraphics[width=0.5\textwidth]{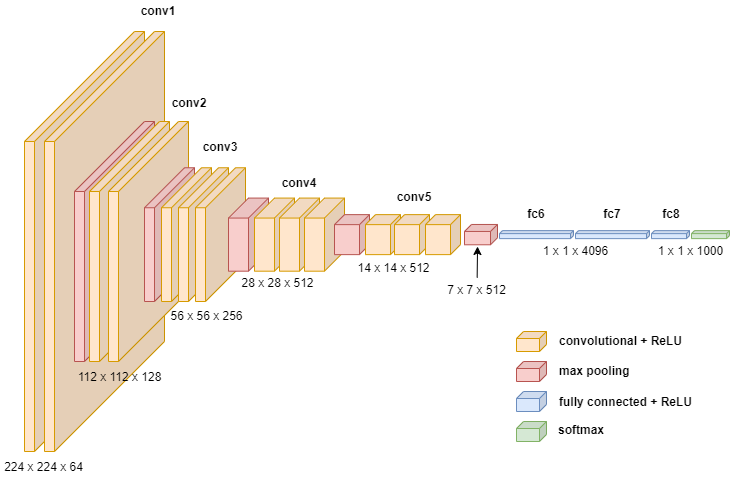}
    \caption{Architecture of baseline CNN in Pytorch.}
    \label{fig:baseline_cnn}
\end{figure}

We begin with an architecture shown in Figure \ref{fig:baseline_cnn}. As a general overview of the model design, the initial layers (leftmost) take in an image of size $224\times 224$, and we use a batch-size of 64.\footnote{Batch-size refers to the number of training examples we use before updating the model weights after backpropagation.} Convolution operations applied across all three colour dimensions repeatedly reduce the size of the image, whilst adding a number of channels (intuitively, this can be thought of as the ``thickening" of the blocks shown in the image). Max pooling also contributes to this downsampling.\footnote{Whilst this report will not go into the details of convolution arithmetic, a fantastic resource for understanding how padding, stride length and convolutions in general perform downsampling can be found \href{https://arxiv.org/pdf/1603.07285.pdf}{here}.} Finally, after several applications of convolution layers, we apply three FC layers, where the inputs are a vector. After applying a softmax to the final FC layer, we output a vector of size 10, representing the probability of the disease being present in the image, for each of the 10 diseases.

Once we have instantiated this architecture, we randomly initialise the weights using Kaiming-He initialisation, where the weights are sampled from a normal distribution following $W \sim \mathcal{N}\left(0, \frac{2}{n^l}\right)$, where $n$ is the size of the input and $l$ is the number of layers. This ensures that we avoid vanishing or exploding gradients by keeping the mean of weights across layers as 0, and the standard deviation as 1 \citep{he2015delving}. We then perform the training process, implemented using a for-loop in Pytorch over batches of training images:
\begin{enumerate}
\item A training image is forward propagated through the CNN, following the convolution, pooling and activations and reaches the final FC-layer, where a softmax activation outputs the predicted probabilities for each class.
\item Th the total loss (using cross-entropy loss) is calculated at the output layer.
\item Backpropagation is used to calculate the gradient of the loss with respect to the model parameters.
\item The gradient is multiplied by the learning rate. This is subtracted from the current weights, giving a new set of parameters that (hopefully) lead to a lower loss and more accurate output predictions.
\end{enumerate}

\subsection{Loss function}
Importantly, we use cross-entropy as our loss function. The loss is simply a function chosen to represent how close (in the output space) the model's predictions are to the desired targets. Here, cross-entropy is chosen because the loss increases as the output probability diverges from the ground truth label for any given image. To see what this means, note that the final FC-layer in our ConvNet has a non-linear softmax activation applied to its outputs. This ensures that all activations are in the range $[0, 1]$, and also that they sum to 1:
\begin{align*}
    \sigma (\mathbf{z})_i = \frac{e^{z_i}}{\sum^K_{j=1} e^{z_j}}
\end{align*}
This gives the formula for the $i$-th element of the softmax-activated input vector, where $K$ is the number of classes (in our case 10). We then apply negative log likelihood to the activated outputs to get the cross-entropy loss. Note how this minimises the distance between the predicted and true probability distributions. This is because cross-entropy loss penalises predictions for being both wrong (i.e. the largest probability is for the wrong class) and confident (i.e. a very high probability for one class, and close to zero for the others). Cross-entropy calculates a separate loss for each class label in a single observation, and takes the sum over labels:
$$ - \sum^K_{x\in C} p(x) \log q(x) $$
Here, $K$ is again the number of classes. $p(x)$ is a binary variable representing whether the particular class label is the true observation value (as a one-hot encoded vector) and $q(x)$ is the predicted probability that the class is the one in the image.

\subsection{Transfer learning and residual networks}
Transfer learning is a powerful tool in using mathematical models in the context of deep learning \citep{zhuang2020comprehensive}. The idea is to instantiate a model with pretrained weights: for example, in computer vision it is common to begin with a model that has been trained for long periods of time on complex datasets such as ImageNet \citep{huh2016makes}. These models have learnt rich representations of images that make them very good for general image detection. The model is then ``fine-tuned" on the specific problem at hand. Transfer learning thus reduces training time, as the model has already learned general computer vision abilities, and only has to learn the minutae of the particular dataset. This was the approach used by \citet{chen2020using} to achieve state-of-the-art results for disease detection on rice-paddies (93\%).

A key part of transfer learning is using architectures that have been iteratively improved over vanilla CNNs. An important adaptation of the CNN was the residual network \citep{he2015deep}, which introduced skip connections to CNNs. The idea was motivated when He et al. noticed that deeper CNNs had poorer performance than exactly the same shallower CNNs, the only difference between the two models being less layers. They proposed that there should exist deep CNNs at least as good as the shallow ones, because extra layers could simply learn the identity mapping after the shallow networks layers had done the work.

\begin{figure}
    \centering
    \includegraphics[width=0.5\textwidth]{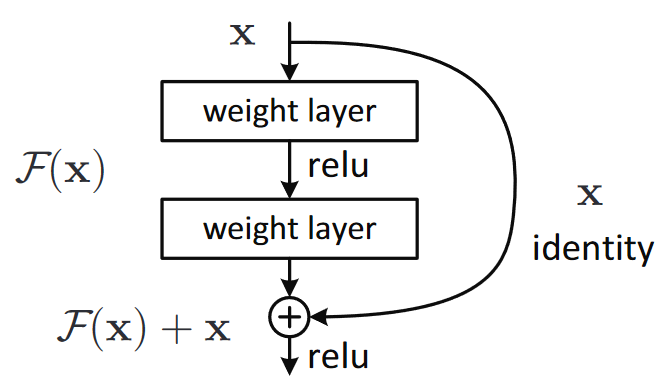}
    \caption{A skip connection in a residual network.}
    \label{fig:skip_connection}
\end{figure}

To allow the network to start from this ``at-least-as-good-as" position, \citet{he2015deep} proposed using skip connections, where every second convolution would also send the pre-convolution activation to the next convolution, thus maintaining information about images that hadn't been downsampled yet. This is shown in Figure \ref{fig:skip_connection}. Note here that zeroing out the weight layers would yield an identity mapping, meaning deeper models could ``copy" the shallow model layers and then maintain performance by setting these weights to zero

In this way, the model learns not to predict certain features, but rather minimise the residual between the input and output, maintaining as much information as possible to pass to the fully-connected layer for prediction. 

\subsection{Baseline}
One of the key parts of this analysis is the sensitivity of performance to model architecture. Hence, several baselines should be established to provide context for improvements driven by different CNNs. Hence, we generate results for two different classes of baseline models:
\begin{enumerate}
    \item Foundational CNN trained with randomly initialised weights
    \item Residual network (ResNet34, a common architecture with 34 layers) with randomly initialised weights
    \item Pretrained ResNet34 that is fine-tuned on current task
\end{enumerate}

The model itself is a CNN, but we will determine how specific model architecture (i.e. ConvNet vs ResNet vs vanilla CNN) affects accuracy, as well as techniques such as transfer learning.

\subsection{Assumptions}
The usefulness of neural networks comes from their capacity for universal function approximation. This negates the need for assumptions required in models such as linear regression, as our network is tasked with simply fitting a function to data. However, there are still several key assumptions we make about the data and models we are using. First, we are assuming that the training and test data come from the same distribution. Without this assumption, any evaluation of the model would be useless, as neural networks are not invariant to shifts in the data distribution. This assumption is justified as the test images come from the same regions of India as the training images. (Saying this, the specificity of the test and training data raises another question about generalisability; see Discussion.) Secondly, we assume that the amount of data is sufficient for the model to learn generalisable trends about rice paddy disease. Again, this assumption seems justified, as we have over 10,000 training images. Finally, we assume that we can select an appropriate capacity of the CNN within the constraints (GPU, memory and training time). A CNN with too few parameters will struggle to learn the complex relationships inherent in the images, and a CNN with too large a capacity has the tendency to overfit. The validity of this assumption will become apparent when evaluating the models below.

\section{Results}
Here, we present the results ordered by complexity of both computer vision techniques and scale of model used. We begin with simple baselines and demonstrating the benefits of transfer learning. We then explore the effect of certain techniques on the performance of the model. After iterating quickly with these techniques on smaller models, we select the most appropriate set of methods and apply them to larger models.

\subsection{Baseline results}
In Table \ref{tab:baselines}, we see that the pre-trained ResNet34 architecture with fine-tuning for 5 epochs achieves similar results to the randomly initialised ConvNet with 50 epochs of training. Immediately, this highlights the benefits of transfer learning, and we conclude that we should use the pretrained ResNet as our base architecture, and iterate on this.
\begin{table}[htbp]
\caption{Baseline model results}
\label{tab:baselines}
\centering
\begin{tabular}{>{\itshape}0l0l0l}\hline
\textup{Model}  & Validation Accuracy & Test Accuracy \\\hline
Randomly initialised ConvNet (10 epochs) & 48.70\% & 47.05\%\\
Randomly initialised ConvNet (50 epochs) & 78.48\% & 79.00\%\\
Randomly initialised ResNet34 (5 epochs) & 79.24\% & 78.51\% \\
\rowcolor{lightgray}
\color{blue}Pretrained ResNet34 (5 epochs) & \color{blue}86.06\% & \color{blue}85.24\% \\\hline
\end{tabular}
\end{table}

% Talk about why the pretrained model is so much better. Obviously the general image recognition thing, but also how using a pretrained model means normalising inputs.

\subsection{Finding the right learning rate}
\begin{figure}
    \centering
    \includegraphics[width=0.5\textwidth]{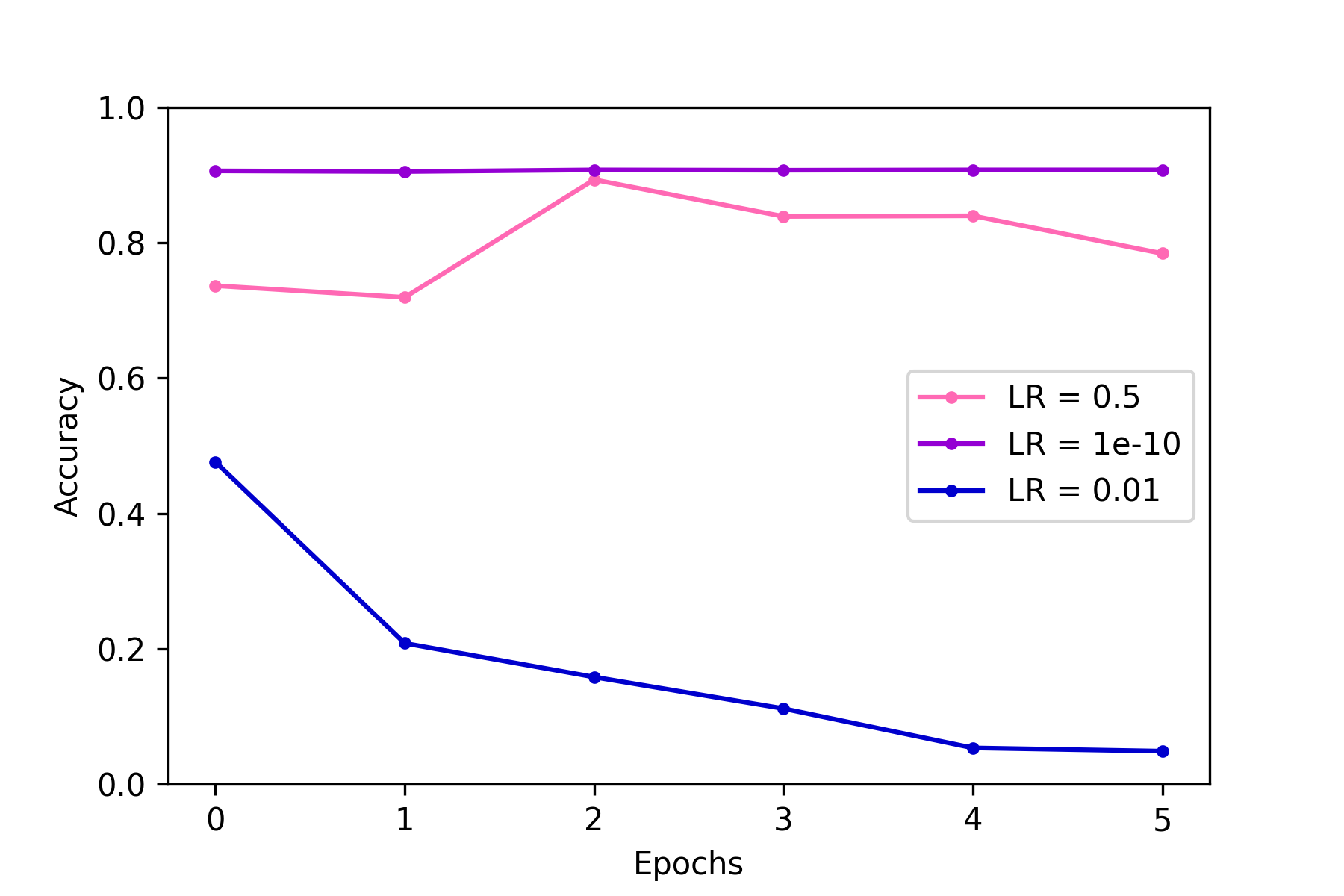}
    \caption{Effects of learning rate on accuracy evolution.}
    \label{fig:lr_effects}
\end{figure}

A key challenge in deep learning has been finding learning rates that are adequately sized. If the learning rate is too high, the loss function will diverge. Conversely, if the learning rate is too low, training times will increase significantly as the model learns at a much slower rate. In Figure \ref{fig:lr_effects}, we show how accuracy evolves over the course of 5 epochs, dependent on different learning rates.

\subsubsection{Cyclical learning rates}
We can improve the learning rate further by using cyclical learning rates. This was proposed in 2015 by Leslie Smith \citep{smith2017cyclical}. The key idea is beginning with an incredibly small learning rate, training for a mini-batch, and then continuously scaling the learning rate up. By tracking the loss as we increase the learning rate, we can use a valley algorithm to determine an optimal learning rate (i.e. a bit before where the loss is minimised on the mini-batch). 

\begin{figure}
    \centering
    \includegraphics[width=0.5\textwidth]{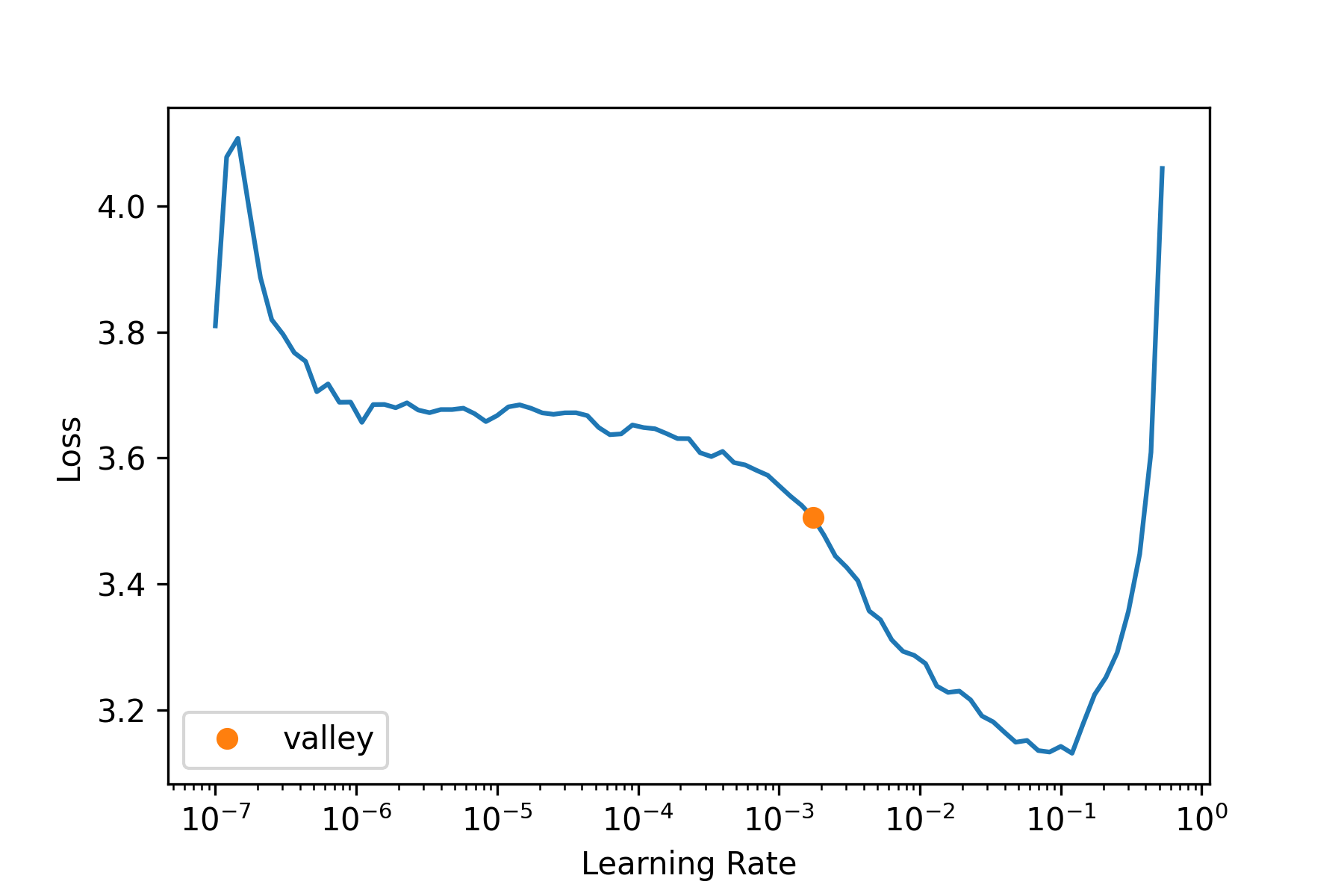}
    \caption{Cyclical learning rate finder.}
    \label{fig:lr_finder}
\end{figure}

Using this idea, we instantiate a ResNet34 model and find that the optimal learning rate is approximately 0.01. Figure \ref{fig:lr_finder} shows this process. After training the model with this learning rate for 5 epochs, the validation accuracy improves to 92.67\%, and the test accuracy improves to 92.08\%.

\subsection{State-of-the-art techniques}
We now conduct a sensitivity analysis of accuracy on a number of widespread computer vision techniques.

\subsubsection{Data augmentation}

\begin{figure}
    \centering
    \includegraphics[width=0.8\textwidth]{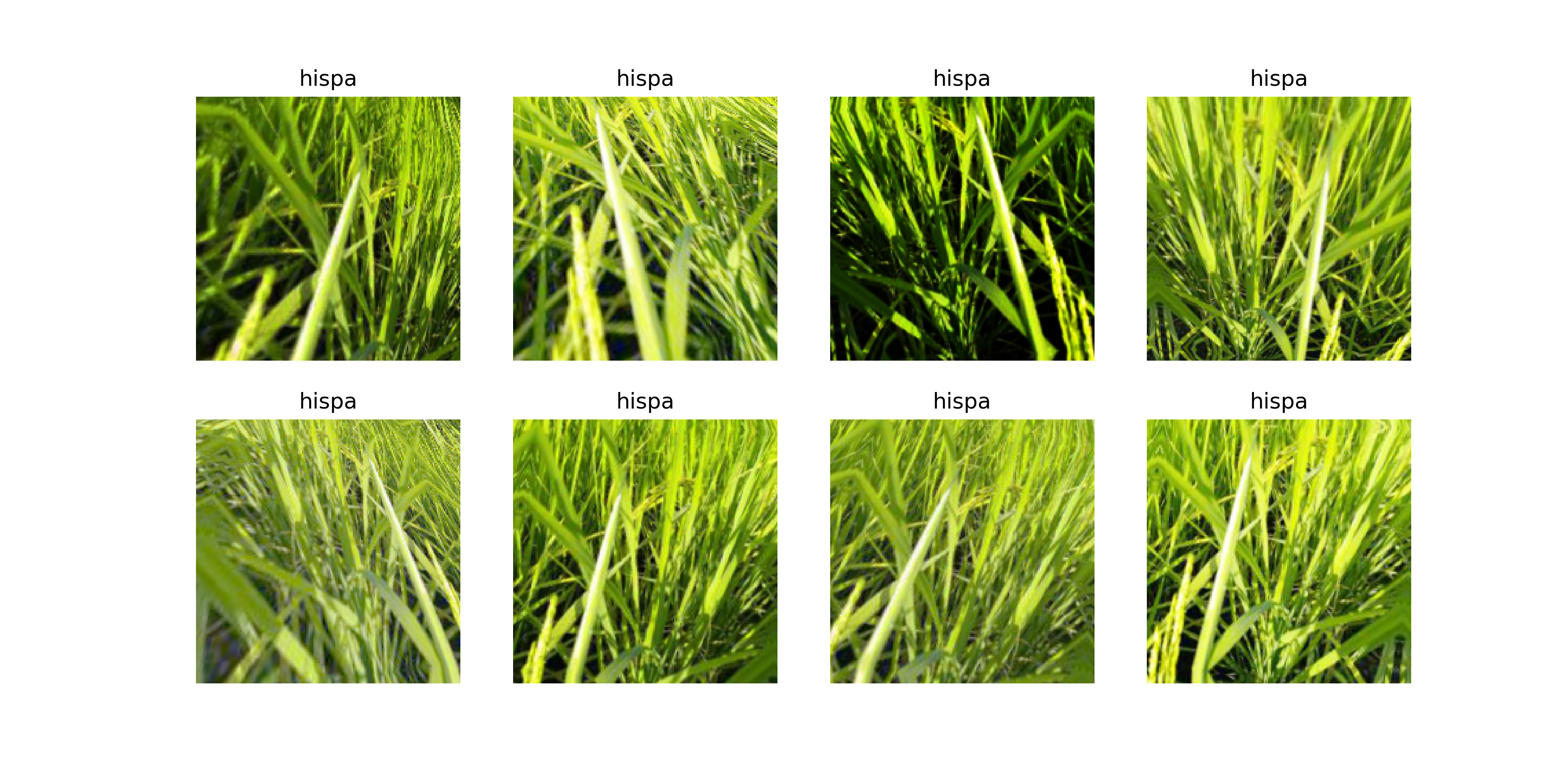}
    \caption{Example of duplicating a training image using data augmentation.}
    \label{fig:data_augmentation}
\end{figure}

To test the effects of data augmentation on model accuracy, we train two models equal in all ways except the augmentations applied to the training images. The first model is trained on images augmented with a variety of techniques, including zoom, horizontal flipping, lighting, contrast, padding, size, stretching, squishing and general affine transformations. Examples of this on a single image are shown in Figure \ref{fig:data_augmentation}. The second model is trained on images that have only been horizontally flipped, with some small contrast and lighting augmentation. 

Interestingly, the second model performs better (93.4\% validation set accuracy vs 91.9\% for the first model). We hypothesise that this is because the main features of each disease are quite similar, particularly in colour, shape and size. Augmenting the images may render some of this information useless. In contrast, other computer vision models (for example, recognising cats and dogs) are invariant under such transformations.

\subsubsection{Progressive resizing}
As mentioned above, a key constraint with our training is timing. One of the biggest factors affecting how long it takes to train a model is the resolution of an image. A key idea to decrease training time is obviously to use smaller image sizes. Above, we iterated with image sizes of $128 \times 128$. However, to improve final accuracy without compromising on training time, we can use a technique called \textit{progressive resizing}: complete most of the training using scaled-down images, and then complete the final few epochs of training using larger images. To ensure the model doesn't get confused when we change image sizes, we freeze the weights of the model body when moving to the larger images and only train the head using fine-tuning, much like we do in transfer learning.

Whilst validation accuracy did not improve (93\%), training time decreased significantly, from approximately 50 seconds per epoch when training using constant image sizes to 35 seconds per epoch using progressive resizing, even when images were scaled up to $224\times 224$. Progressive resizing can also be seen as a form of data augmentation, which can improve generalisation.

\subsubsection{Test-time augmentation}
We have been using random cropping as a way to get some useful data augmentation, which leads to better generalization, and results in a need for less training data. However, at validation time, our random cropping has been selecting the maximal centre square of the image. This can cause issues, as the model may miss out on key parts of the rice paddy that signal certain aspects of disease, which will have detrimental effects on accuracy. Whilst many solutions have been proposed to this problem, such as squishing or stretching the image to fit the square pixel count, these introduce new problems, as the model may not be invariant to such affine transformations.

Instead, we use \textit{test-time augmentation} (TTA), where each time we validate an image, we create multiple copies of it to be predicted. This involves not just cropping, but any version of augmentation we passed in during training. We then take an average of predictions across image versions, often resulting greatly improved accuracy.

Starting from a model with accuracy of 92.8\%, using TTA in validation considerably decreases the error rate, giving a validation accuracy of 95.2\%.

\subsubsection{Mixup}
Mixup provides an additional form of augmentation by training the CNN on linear combinations of examples from different classes \citep{zhang2017mixup}. It allows non-expert dependent augmentation that can be scaled with the requirements of the dataset. To use Mixup, we select two random images from the training dataset, along with a random weight parameter $\lambda$. By taking a weighted average of the two images, we produce a new training image $\tilde{\mathbf{x}}$ (independent variable); similarly, the same weighted average of the image labels gives the new label $\tilde{\mathbf{y}}$  (dependent variable).
\begin{align*}
    \tilde{\mathbf{x}} &= \lambda \mathbf{x}_i + (1-\lambda)\mathbf{x}_j \\
    \tilde{\mathbf{y}} &= \lambda \mathbf{y}_i + (1-\lambda)\mathbf{y}_j
\end{align*}

However, introducing Mixup makes the classification task significantly harder: the model is now required to predict two classes per image, as well as the weighting applied to each one. Thus, Mixup usually requires significantly more epochs to get better results. Indeed, we show this after training the ResNet34 for 25 epochs; without Mixup, the model overfits and achieves an accuracy of 93.8\%. However, with Mixup the validation accuracy improves to 97.3\%.

\subsubsection{ConvNext small}
From above, we concluded that minimal data augmentation is best (using just horizontal flipping and small lighting exposure changes). We also learned that TTA can significantly improve accuracy at inference time, and that mixup only provided improvements when training for 25+ epochs. Taking these ideas, we now apply them to a larger model. Training for 5 epochs with appropriate augmentation and TTA, the model achieves an accuracy of 95.58\% on the validation set and 95.67\% on the test set.

\subsection{Larger architectures}
Evidently, models with more parameters can discover more complex and general patterns that underlie the relationships in the data distribution. However, increased capacity also increases the propensity to overfit; the model can become large enough to start ``memorising" training data. 

Furthermore, larger models require larger amounts of GPU RAM. That is, we must use smaller batch sizes of image data to ensure we don't run out of memory on the GPU. The GPUs used for this analysis have 16280MiB of memory, which isn't much compared to research setups, which typically have 24-128 GiB of GPU memory. We address our memory concerns in three ways: image sizing, gradient accumulation, and mixed-precision training. We have already noted the use of $128\times 128$ sized images (rather than $224\times 224$) to reduce training times. We discuss the two other techniques here.

\subsubsection{Gradient accumulation}
Gradient accumulation reduces memory use through a clever innovation: model weights are accumulated for several batches of training, and then the weights are updated all at once, rather than after each batch. The training loop remains the same, but the memory use is reduced by $\alpha$, the accumulation parameter (representing how many batches to accumulate gradients for). The only difference between gradient accumulation and regular training is dividing the batch-size by $\alpha$.

We use one class of disease to quickly train a selection of different architectures to determine which ones will fit in memory. We begin with an accumulation factor of 1, and double it each time we exceed the 16 MB limit. To find how much memory each model is using, we can use CUDA to list current GPU processes and subsequently the memory. A list of model memory usages is shown in Table \ref{tab:memory}. Note how increasing $\alpha$ for ConvNext small decreases GPU memory usage considerably.

\begin{table}[htbp]
\caption{Model memory usage}
\label{tab:memory}
\centering
\begin{tabular}{>{\itshape}0l0l0l}\hline
\textup{Model}  & Accumulation factor $\alpha$ & GPU Memory (MB) \\\hline
ConvNext Small & 1 & 4173 \\
ConvNext Small & 2 & 3101 \\
ConvNext Small & 4 & 2585 \\
ConvNext Large & 2 & 11023 \\
Vit Large & 2 & 15343 \\
Swin V2 & 2 & 13499 \\
\hline
\end{tabular}
\end{table}

The models were chosen from a commonly available \href{https://github.com/rwightman/pytorch-image-models}{list} of pretrained image models implemented in Pytorch as part of the standard repository. Using this technique, it appears that ConvNext, Swin and Vit models (both small and large versions) all fit in memory.

\subsubsection{Mixed-precision training}
Another way to reduce training time and memory usage is to explore how floating point accuracy affects these two constraints. Deep learning packages typically represent all computations in single-precision, using 32-bit floats (FP32) to represent inputs and weights. To reduce data usage, many deep learning practitioners have attempted to train models in half precision, using 16 bits to represent numbers (FP16), which should use half the RAM and hence allow double the batch size. However, this obviously represents a trade-off with accuracy, as by definition FP16 is less precise in representing model parameters.

To address this, \citet{micikevicius2017mixed} introduced \textit{mixed-precision training}, which uses FP32 and FP16 interchangeably depending on what stage of training we are up to. For instance, forward propagation and backpropagation (gradient computations) are done in half-precision, but the weights update itself is done in full-precision. This is largely because small learning rates mean the updates to weights will be small, which can only be represented accurately in full precision. To achieve this, we store a copy of the weights as we train in full 32-bit precision.

On a NVIDIA GPU, mixed-precision training is easily implementable, so we apply it to the training of the following large models.

\subsubsection{ConvNext large}
Using these techniques for reducing memory, we now train a ConvNext large model using larger-resolution images, along with the techniques we selected for ConvNext small, for 5 epochs. With this, we achieve an accuracy of 97.69\% on the validation dataset, and 97.89\% on the test dataset, improving a few percentage points over ConvNext small.

\subsection{Ensemble}
Using ensembles of fine-tuned computer vision models is common in the literature \citep{qummar2019deep}. The idea here is that different models initialised with different random seeds will make uncorrelated errors when predicting unseen data. If the predictions are averaged, the expected value of these uncorrelated errors is zero. Whilst errors won't be completely uncorrelated in practice, we can still leverage this idea to create an ensemble of models.

\begin{figure}
    \centering
    \includegraphics[width=0.5\textwidth]{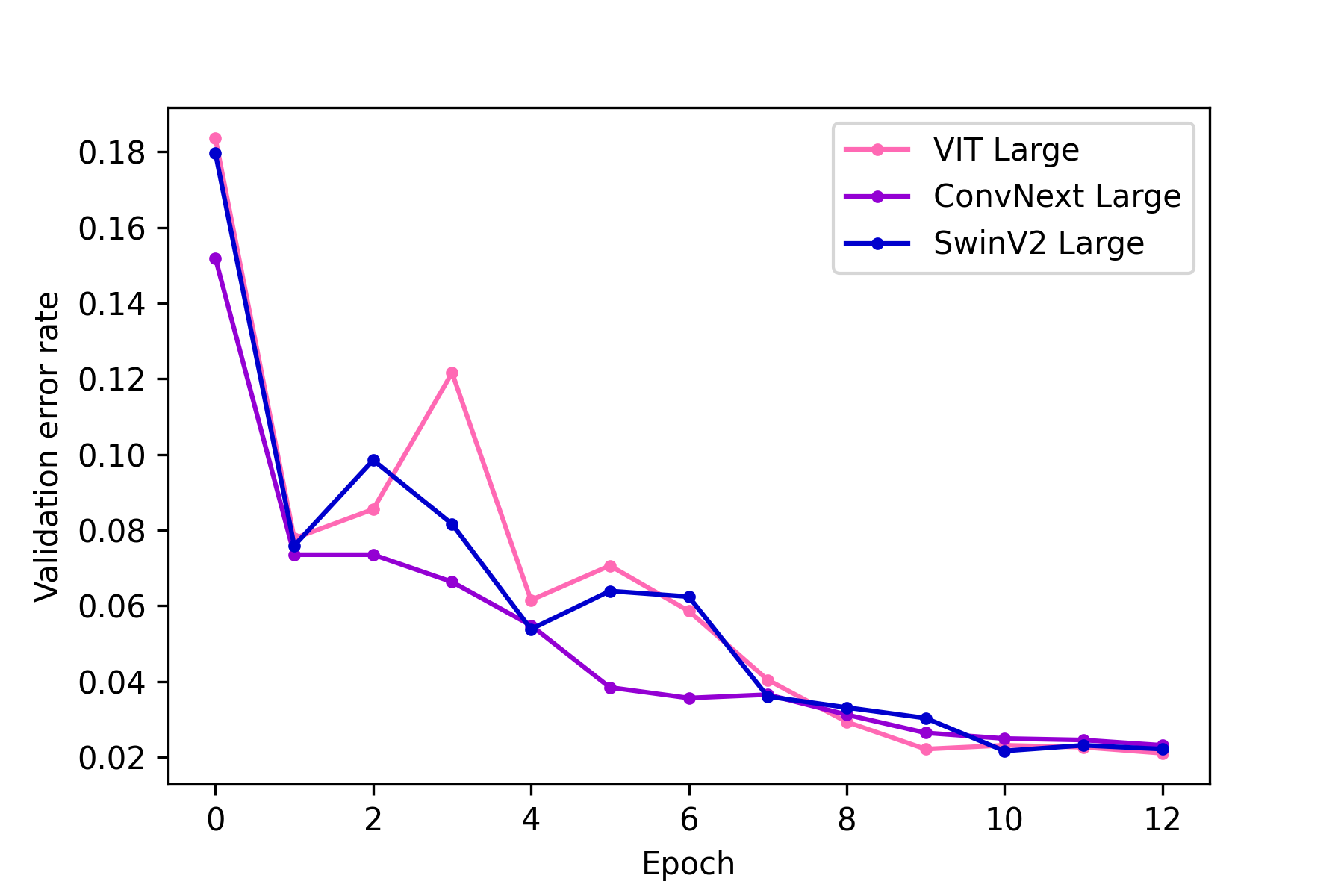}
    \caption{Error rates over the course of training for the ensemble component models.}
    \label{fig:ensemble}
\end{figure}

The error rate for each of the models in the ensemble is shown in Figure \ref{fig:ensemble}. To create an ensemble model, we simply take the average of predictions across all models. During training, it was noticed that VIT performed slightly better than other models, so VIT was weighted double in the prediction averages. This yielded an accuracy of 98.90\% on the validation set, and 98.69\% on the test set, by far the most accurate model.

% TO-DO: insert some stuff mentioning why I've done error rate instead of accuracy here

\subsection{Final results}
In Table \ref{tab:final}, we present the final model results for the most relevant architectures and techniques. Clearly, ensemble learning with ConvNext, Vit and Swin models achieves the best accuracy. 
\begin{table}[htbp]
\caption{Final model results}
\label{tab:final}
\centering
\rowcolors{5}{white}{lightgray} 
\begin{tabular}{>{\itshape}0l0l0l}\hline
\textup{Model}  & Validation Accuracy & Test Accuracy \\\hline
Fine-tuned ResNet34 & 86.06\% & 85.24\%\\
Fine-tuned ResNet34 with selected techniques (5 epochs) & 95.20\% & 95.44\%\\
Fine-tuned ConvNext Small with selected techniques (5 epochs) & 95.58\% & 95.67\% \\
Fine-tuned ConvNext Large with selected techniques (5 epochs) & 97.69\% & 97.89\% \\
\color{blue}Ensemble (ConvNext, Vit and Swin) (12 epochs) & \color{blue}98.90\% & \color{blue}98.69\% \\\hline
\end{tabular}
\end{table}

\section{Discussion}
The focus of this mathematical modelling was to conduct a sensitivity analysis of how both model architecture and computer vision techniques affect accuracy when predicting rice paddy disease. Additionally, to ensure results weren't driven by computation power alone, we set the constraint of training the predictive models with only one GPU with 16 GB of RAM and short training times (less than five minutes per epoch).

When comparing to vanilla CNNs, model architecture yields the biggest increase in performance. For instance, a CNN initialised with random weights required 50 epochs to achieve 80\% accuracy, whereas a randomly initialised residual network implementing skip connections in the CNN only required 5 epochs to get to the same point.

However, once this initial improvement was made, further changes to architecture achieved minimal improvement. For instance, using a ConvNext architecture (which is currently considered the state-of-the-art CNN for computer vision) over a ResNet34 yielded similar accuracies. The next jump in accuracy was achieved by using transfer learning, where we instantiate a model with pretrained weights and fine-tune it on the paddy disease task. Doing so improved accuracy by approximately 10\%. This improvement of pretrained models over the baseline networks demonstrates the power of transfer learning, and this allowed us to select only pretrained models to iterate on.

We then explored how accuracy was sensitive to computer vision techniques that weren't dependent on architecture. We found that using appropriate learning rates, data augmentation and test-time augmentation yielded further improvements in accuracy. Some techniques such as Mixup required many epochs of training to yield improvements, and as such were disqualified from use due to our self-imposed training time constraints.

A key lens through which we viewed our analysis was a meta-overview of techniques that allowed us to train faster and with minimal GPUs and memory. Whilst image downsizing was implemented from the beginning, other useful techniques included gradient accumulation and mixed precision training. Progressive resizing proved effective in reducing training times, but resulted in considerable degradation in model accuracy, requiring more epochs to correct.

% Talk about more effective validation methods (i.e. cross-fold validation)

% Talk about how we could have used AUC as a metric as well

% Probably do some interpretability (CAM CNN learner, what each layer is looking at). Talk about Zeiler and Fergus
% As part of this look at a confusion matrix

There are several potential limitations with this work. The first is potential generalisation problems. The dataset came entirely from paddy fields in the Tirunelveli district of Tamilnadu, India. The training data distribution may thus be significantly different from general crop images. For instance, bacterial blight is a paddy disease that requires humidity above 70\%, and thus may be absent in certain types of fields \citep{mew1987current}. However, \citet{temniranrat2021system} note that most rice crops look similar, and climates suitable for rice are specific enough that the same diseases manifest themselves in most locations. An additional limitation, largely constrained by minimising training time, was that cross-validation was not used. Here, we evaluated model performance using a validation set and a further hold-out test set (ensuring hyperparameters could not be optimised to this specific distribution). Whilst this is effective, an even better approach would be to use $k$-fold cross validation, where the data is divided into $k$ equal subsets, $k-1$ of which are used for training and the remaining for validation. Repeating this process can yield statistical insights into model accuracy (i.e. standard deviation as well as mean). However, given the size of the dataset, this would be unlikely to lead to significant generalisation improvements.

Saying this, the above limitations also engender future research directions. For instance, a further dataset from a different region with varied climate could be constructed to test the generalisability of the trained model. Additionally, work could be done on understanding model predictions, particularly how each layer decomposes each image. There are several tools which may allow this for CNNs specifically, such as class activation maps to visualise which pixels activate which layers of the network. Finally, work could be done exploring how metrics other than accuracy, such as ROC AUC (receiver-operating characteristic, area-under-curve) can lead to improved interpretability and better model performance.

\section{Conclusion}
In this work, we leveraged the power of deep learning to achieve state-of-the-art results for rice paddy disease detection, improving accuracy from a literature-best of 93\% on five classes of disease to 98.7\% on a more difficult task of predicting ten diseases. We did this within the constraints of one GPU, limited RAM and reduced training times, which isolated the effects on performance of (i) model architecture and (ii) computer vision techniques. Specifically, we conducted a sensitivity analysis of how these two mathematical modelling considerations contributed to classification accuracy. We found that model architecture led to small improvements, but it was the techniques that drove the best accuracy. Specifically, finding an appropriate learning rate allowed the model to converge quicker. Using test-time augmentation and experimenting with appropriate data augmentation also yielded significant improvements. Further, monitoring GPU memory usage and implementing gradient accumulation allowed us to ensemble relatively large models, despite limited RAM, and ensured all training runs were under 12 epochs and 5 minutes per epoch. This highlights how iterative modelling can beat academic benchmarks, despite reduced modelling capacities. This is particularly important in the current problem: any model that predicts rice paddy disease will likely have to be light-weight enough to be implemented on a mobile device in order to be usable by rural farmers, ruling out excessively large architectures.

\bibliography{references}
\bibliographystyle{apalike}

\end{document}